\documentclass[journal]{IEEEtran}

\usepackage{cite}

\usepackage{times}
\usepackage{epsfig}
\usepackage{graphicx}
\usepackage{amsmath}
\usepackage{amssymb}
\usepackage{mathtools}
\usepackage{makecell}
\usepackage{booktabs}
\usepackage{verbatimbox}
\usepackage[pagebackref=true,breaklinks=true,letterpaper=true,colorlinks,bookmarks=false]{hyperref}
\usepackage[font=small,labelfont=bf]{caption}

\ifCLASSINFOpdf

\else

\fi

\usepackage{pifont}

\begin{document}

\title{Model Pruning Based on Quantified Similarity of Feature Maps}

\author{Zidu~Wang,~\IEEEmembership{}
        Xuexin~Liu,~\IEEEmembership{}
        Long~Huang,~\IEEEmembership{}
        Yunqing~Chen,~\IEEEmembership{}
        Yufei~Zhang,~\IEEEmembership{}
        Zhikang~Lin,~\IEEEmembership{}
        \\ and~Rui~Wang,~\IEEEmembership{Senior Member,~IEEE}

\thanks{This work was supported by the National Natural Science Foundation of China under Grant No. 62173158 and 61379134. (Corresponding author: Rui Wang.)}
\thanks{Zidu Wang and Xuexin Liu are with the School of Computer and Communication Engineering, University of Science and Technology Beijing, Beijing 100083, China, and also with the Institute of Automation, Chinese Academy of Sciences, Beijing 100190, China and University of Chinese Academy of Sciences, Beijing 100190, China (e-mail: wangzidu2022@ia.ac.cn; liuxuexin2022@ia.ac.cn).}
\thanks{Long Huang, Yunqing Chen, Yufei Zhang, Zhikang Lin and Rui Wang are with the School of Computer and Communication Engineering, University of Science and Technology Beijing, Beijing 100083, China (e-mail: 2964901878@qq.com; serein7z@163.com; zyf18610299509@126.com; linjy15318955887@163.com; wangrui@ustb.edu.cn).}

}

\maketitle

\begin{abstract}

Convolutional Neural Networks (CNNs) has been applied in numerous Internet of Things (IoT) devices for multifarious downstream tasks. However, with the increasing amount of data on edge devices, CNNs can hardly complete some tasks in time with limited computing  and storage resources. Recently, filter pruning has been regarded as an effective technique to compress and accelerate CNNs, but existing methods rarely prune CNNs from the perspective of compressing high-dimensional tensors. In this paper, we propose a novel theory to find redundant information in three-dimensional tensors, namely Quantified Similarity between Feature Maps (QSFM), and utilize this theory to guide the filter pruning procedure. We perform QSFM on datasets (CIFAR-10, CIFAR-100 and ILSVRC-12) and edge devices, demonstrate that the proposed method can find the redundant information in the neural networks effectively with comparable compression and tolerable drop of accuracy. Without any fine-tuning operation, QSFM can compress ResNet-56 on CIFAR-10 significantly (48.7\% FLOPs and 57.9\% parameters are reduced) with only a loss of 0.54\% in the top-1 accuracy. For the practical application of edge devices, QSFM can accelerate MobileNet-V2 inference speed by 1.53 times with only a loss of 1.23\% in the ILSVRC-12 top-1 accuracy.

\end{abstract}


\IEEEpeerreviewmaketitle

\begin{figure}[t]
\begin{center}
   \includegraphics[width=0.9\linewidth]{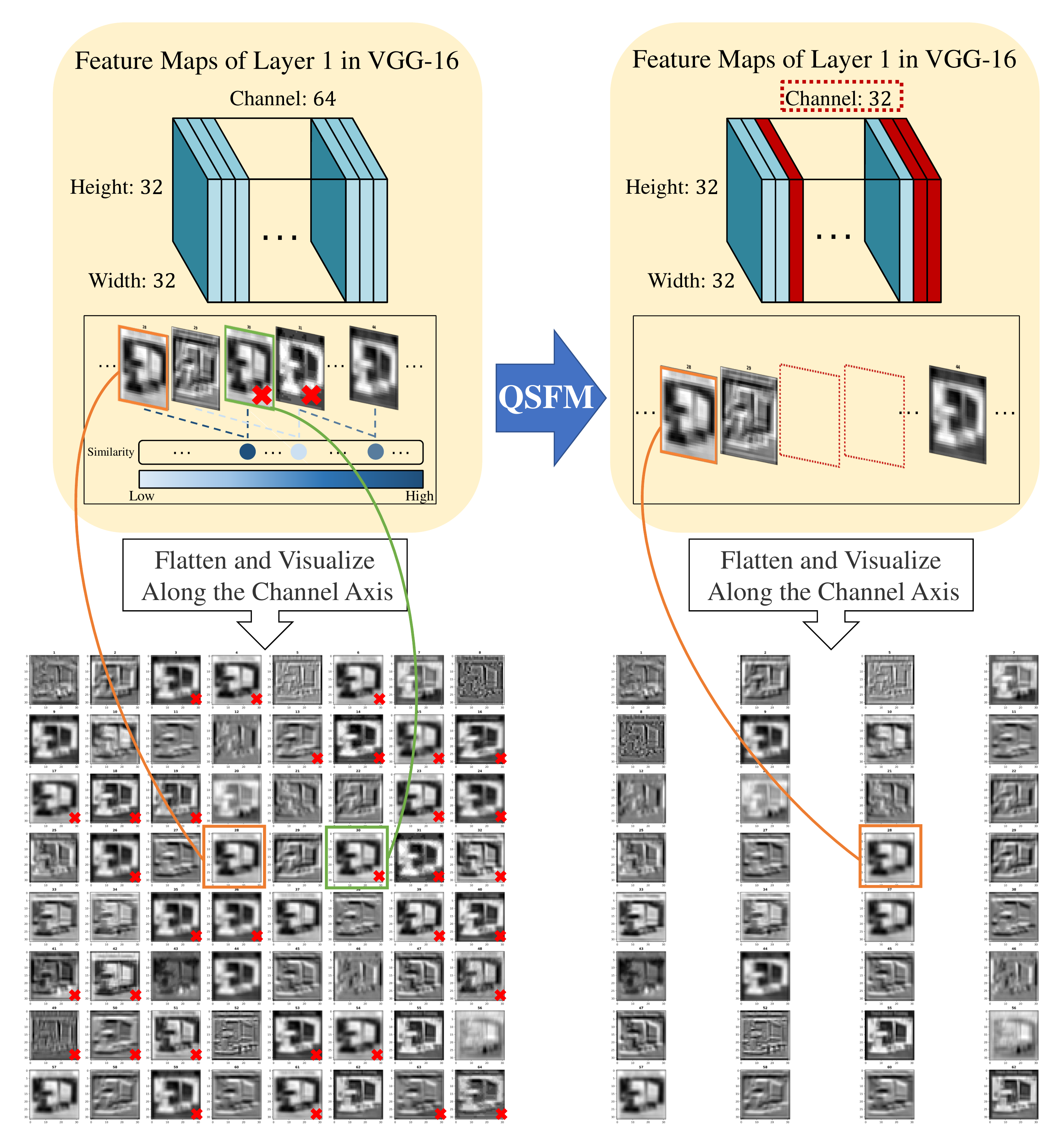}
\end{center}

\captionsetup{font={footnotesize}} 
\vspace{-0.4cm}
   \caption{Channels deleted after QSFM are marked in red. The upper part of the figure shows that QSFM prune model by deleting similar feature maps, for example, the feature map with orange box is kept, while the similar one with green box is deleted.  The lower part visualizes the feature maps from the first layer of VGG-16.}
\vspace{-0.5cm}
\label{wzd1}
\end{figure}

\section{Introduction}

\IEEEPARstart{W}{ith} the popularity of the Internet of things (IoT), widely distributed mobile and IoT devices generate more data, which may be more than that generated by large cloud data centers. Some IoT applications may require a short response time, some may involve private data, and some may generate a large amount of data, which may cause a heavy burden on the network \cite{Shi2016}. Therefore, it need to be more effective for processing data at the edge of topological networks. Artificial Intelligence (AI) has demonstrated its great ability in data processing and analysis \cite{DBLPiotj}, many edge devices also load CNNs \cite{Deng2020}, such as VGGNet \cite{Simonyan2015}, ResNet \cite{He2016}, GoogLeNet \cite{Szegedy2015}, MobileNet \cite{Sandler2018} and DenseNet \cite{Huang2017} to meet the needs of different tasks like image classification \cite{He2016,Szegedy2015}, object detection \cite{Ren2015,TanPL20}, 3D reconstruction \cite{GuoZYYLL20,YuSFR21} and so on.

	However, with the improvement of large CNNs performance, there are many problems, such as huge amouts of parameters, terrible computing consumption and large memory requirements which become the challenges for edge and mobile embedded devices. 

	Compressing and accelerating neural network models is a hot topic recently. At present, the typical works include network quantization \cite{Guo2017, Hu2018}, low-rank approximation \cite{Masana2017}, weight sharing \cite{Wang2017, Chen2016, Son2018} and weight pruning \cite{Lee2019, Yang2019}. These methods need specific hardware or software libraries to run, and some of them can not solve the problems described above comprehensively. There is another lightweight technique defined as filter pruning \cite{Li2017, Lin2020, He2017, He2018, Lin2019, Zhao2019}, also known as coarse-grained pruning. By comparison, filter pruning is not restricted to special hardware or software, and suitable as well as universal for CNNs in various types of tasks. Tab.~\ref{table-0} lists the attributes existing in relevant methods, and it can be seen that filter pruning is more universal among them.

\begin{table}
\captionsetup{font={footnotesize}} 
\caption{Attributes in Existing Model Compression Methods.}
{ \ding{51}or \ding{55} represents yes or no. * denotes that the corresponding method can solve the problem in specific situations. }
\begin{center}
\addvbuffer[-5pt -20pt]{
\setlength{\tabcolsep}{1mm}{
\begin{tabular}{c|c|c|c|c|c}
\toprule
\scriptsize{Attributes} & \thead{\scriptsize{Weight}\\\scriptsize{Pruning}} & \thead{\scriptsize{Network}\\\scriptsize{Quantization}} &  \thead{\scriptsize{Weight}\\\scriptsize{Sharing}} & \thead{\scriptsize{Low-Rrank}\\\scriptsize{Approximation}}&\thead{\scriptsize{Filter}\\\scriptsize{Pruning}}\\
\hline\midrule
\thead{\scriptsize{Network Structure}\\\scriptsize{Maintenance} }&\ding{55} &\ding{51}  &  \ding{51}& \ding{55}   & \ding{51}\\[-1ex]
\thead{\scriptsize{Not Required Special}\\\scriptsize{Compilation Library}} &\ding{55} & \ding{55} &  \ding{51}&  \ding{51}  &\ding{51}\\[-1ex]
\thead{\scriptsize{Robustness in}\\\scriptsize{Various Hardware}}&\ding{55} &\ding{55}  &  \ding{51}&  \ding{51}  & \ding{51}\\[-1ex]
\thead{\scriptsize{Acceleration}\\\scriptsize{for Inference}} &* & *  &  \ding{55}&  \ding{51}  &\ding{51}\\[-0.7ex]
\bottomrule
\end{tabular}}}
\end{center}
\label{table-0}
\end{table}

	Filter pruning has been shown to be significant in network slimming \cite{Frankle2019, Malach2020}. Great progress has been made in filter pruning, but there are also some problems, such as incurring extra hyper-parameters, destroying model structure and remaining redundant information. \cite{He2018, Lin2019, Lin2020a} put forward specific optimization objectives and constraints, and utilize heuristic optimization algorithms to train CNNs jointly. These will bring a new set of hyper-parameter problems. That means, aiming to deploy strategies, extra tricks to further adjust hyperparameters of heuristic algorithms are needed, which is not flexible for practical application. Moreover, jointed training, combined with sparsity regularization penalty, will destroy the structure of the model itself. In \cite{Huang2018, He2017, Zhao2019, Liu2017, Yu2018, Lin2020, Li2017, cuili, 9552245}, filters are pruned by concrete regulations, which are explicit in generating filter importance. These methods need to sort the importance from high to low and then delete the low importance filters without manual adjustment. \cite{9552245} compress CNNs with pruning and tucker tensor decomposition, which destroys the original model structure and complicates deployment. Although \cite{Yu2018, He2017} have achieved success in compressing models, they rely too much on intuition and lack basic theoretical guidance. Meanwhile, other rule-based methods provide sufficient theoretical support to find out this redundant information, but still remain defects. \cite{Li2017} applies the L1 norm to filters as the evaluation criterion of importance. However, all layers used the same trim scale, which brought suboptimization. \cite{Huang2018, Liu2017} use scaling factor to measure the importance of parameters,  pay too much attention to the parameters themselves and ignore information redundancy in the outputs generated by these parameters. \cite{Lin2020} proposes a method named Hrank to prune filters with low-rank feature maps. Similar high-rank feature maps may also contain redundant information, but they have not been removed.

    We believe that the reason for the shortcomings of the filter pruning methods mentioned above is that the relationship between model pruning and tensor compression is not fully utilized. In CNNs, 2D feature maps output by the middle layer constitute 3D tensors. As shown in Fig.~\ref{wzd1}, it can be found that the original model contains many similar feature maps by visualizing the feature maps output by the first layer of VGG-16. Inspired by the similarity between feature maps, we introduce the similarity function to quantify the similarity between feature maps and delete similar feature maps to prune the CNNs. The whole process is analogous to constructing the maximum linearly independent group of 3D tensors, as shown in the upper part of Fig.~\ref{wzd1}. GhostNet\cite{Han2020} explains and applies the similarity of feature maps, but it makes use of this property from the perspective of constructing the structure of convolutional neural networks, and does not perform an in-depth investigation about how to use this property to model pruning, which is different from the tensor compression and CNNs pruning in this paper.

	Our main contributions are as follows:

\begin{itemize}

\item We propose a theory to find out the redundant information in 3D tensors by quantifying the similarity between any two feature maps, namely QSFM. QSFM can compress 3D tensors in CNNs to prune models.

\item Based on QSFM, we prune a variety of convolution layers such as common convolution layers and depth-wise separable convolution layers, which do not need special software library and can accelerate the model inference speed.

\item Experiments demonstrate that our method is applicable to almost all CNNs, such as VGGNet, MobileNet and ResNet. On CIFAR-10 \cite{CIFAR10}, CIFAR-100 \cite{CIFAR10} and ILSVRC-12 \cite{imagenet},  QSFM shows superior performance over existing prior filter pruning methods. QSFM can improve the inference speed of CNNs in practical applications of mobile and edge devices.

\end{itemize}

    The remainder of this paper is organized as follows. Section II discusses the recent achievements related to deep learning model compression. Section III presents QSFM compress 3D tensors by taking advantage of the similarity between feature maps. Section IV conducts experiments and analyses to verify the feasibility of QSFM on different model structures, multiple datasets and different devices. Conclusions are given in the Section V.

\section{Related Work}

According to existing popular model compression methods, three research domains are most relevant to our approach, which can  be categorized into network quantization, weight pruning and filter pruning.

	\textbf{Network Quantization}. \cite{Guo2017} proposed the network sketching method, which used the convolution with binary weight sharing: for convolution calculation with the same input, the result of the previous convolution is retained, and the same part of the convolution filters directly multiplex the result. \cite{Hu2018} projected data into Hamming space by hashing, and transformed the problem of learning binary parameters into a hashing problem under inner product similarity. Different from the traditional 1-Valued or weighted mean, \cite{Zhu2017} proposed trained ternal quantification (TTQ), which used two trainable full precision scaling coefficients to quantify the weight to $ \{-w_n,0,w_p \}$, and asymmetric weights made the network more flexible. These network quantization methods can accelerate the inference speed by specific library (denoted as * in Tab.~\ref{table-0}) such as \cite{fbgemm}, while only utilizing the network quantization algorithm without the assistance of other libraries may not able to accelerate speed in reality. Unlike our approach, these methods aim to quantify the weight of filters into discrete values, while our approach is to reduce channels for the 3D tensors of convolutional layer output. Besides, these methods obtain a very high level of compression in saving storage space and significantly accelerate inference, but extreme compression (continuous to discrete) may limit the fitting ability of model and may bring a considerable loss of accuracy, damage the model performance. Additionally, the value of the possible weight network needs to be further explored.

	\textbf{Weight Pruning}. Weight pruning can remove any redundant parameter or redundant connection of the expected proportion of the network without restriction, but it will bring the problem of irregular network structure, making it difficult to effectively accelerate after pruning. Recently, \cite{Lee2019} proposed SNIP method that the importance of the connection was determined by sampling the training set several times in the initialization phase of the model, and the pruning template was generated in the meantime. After training, there is no need to iterate pruning and fine-tuning by the alternating cycle process. \cite{Yang2019} used weighted sparse projection and input masking to provide quantifiable energy consumption, taking energy consumption budget as the optimization constraint of network training, and the sparse network can be obtained by utilizing dynamic pruning method which can recover the important connections removed by mistake. Due to requirement of special sparse matrix operation library and hardware, it is also not convenient and universal to use weight pruning to implement acceleration for model.

\begin{figure}[t]
\begin{center}
   \includegraphics[width=0.9\linewidth]{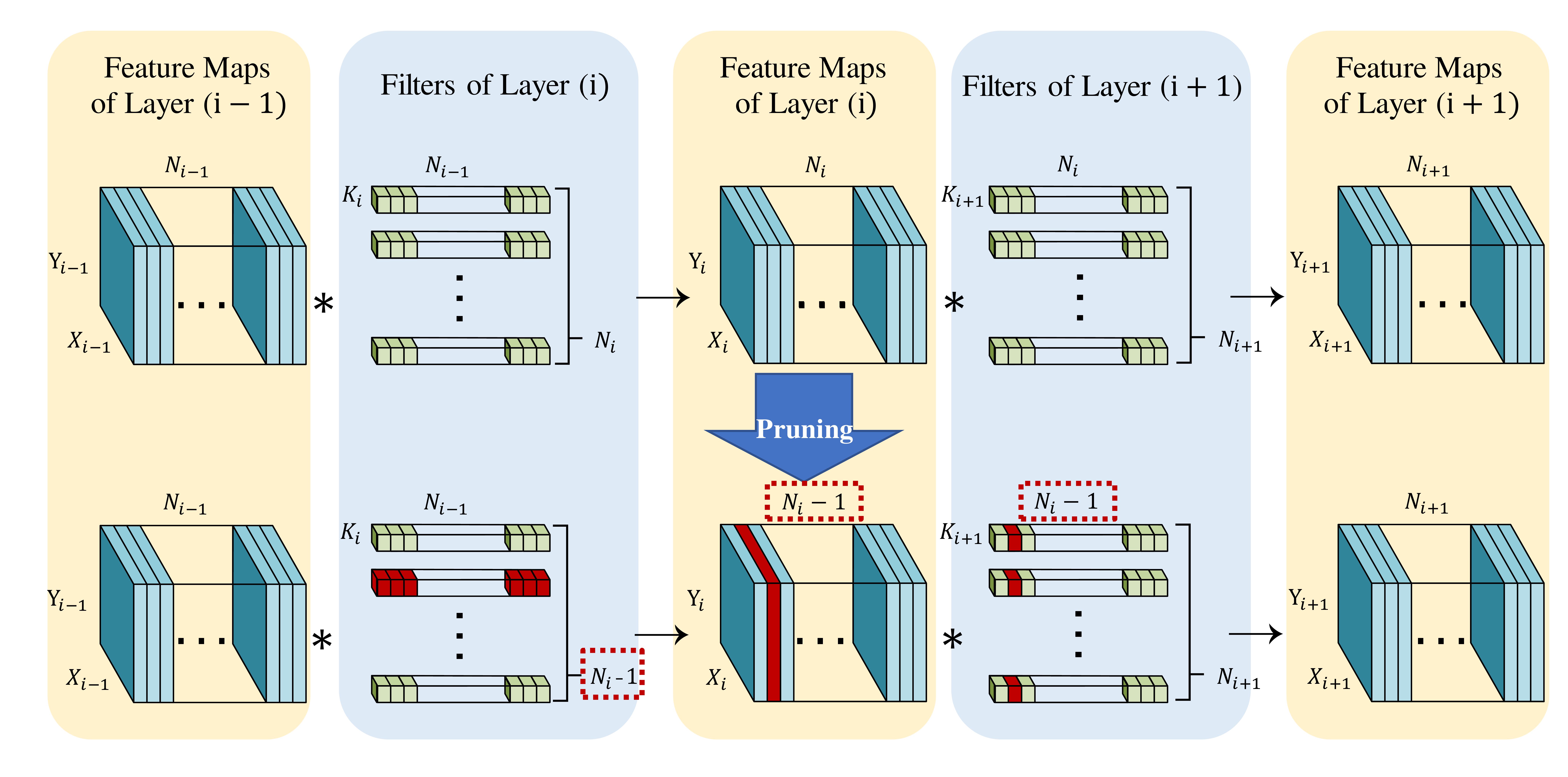}
\end{center}
\captionsetup{font={footnotesize}} 
\vspace{-0.4cm}
   \caption{Overview of filter pruning process. The figure shows the change after deleting the second channel of Layer(i) output. The red parts indicate that they have been deleted, and the values enclosed by the red dotted line indicate that the dimensions have been changed. '*' means convolution operation.}
\vspace{-0.4cm}
\label{fig2}
\end{figure}

	\textbf{Filter Pruning}. Also named as coarse-grained pruning, it considers filters as the minimum prunning units, which can make the network 'narrow' and can directly achieve effective acceleration on existing software or hardware. Fig.~\ref{fig2} shows how filter pruning reduces the output channels of the convolution layer. Compared with human-crafted pruning, \cite{He2018} made the model compression completely automatic and performed better. It utilized DDPG (Deep Deterministic Policy Gradient) as the controller to generate the specific compression ratio in continuous space. Different from the previous hard pruning and label dependent pruning methods, \cite{Lin2019} proposed a label free Generative Adversarial Learning (GAL) method, which used sparse soft mask pruning network to scale the output of specific structure to zero. It learnt pruning networks with sparse soft masks in an end-to-end manner. Due to heuristic optimization algorithm and jointed training, \cite{He2018, Lin2019} will bring a new set of hyper parameter problems. \cite{He2017} proposed an iterative two-step algorithm to effectively prune each layer, by a Least Absolute Shrinkage and Selection Operator (LASSO) regression based channel selection and least square reconstruction. \cite{Zhao2019} proposed a variational Bayesian scheme for pruning convolutional neural networks in channel level, introducing a stochastic variational inference to estimate the distribution of channel saliency induced by a sparse prior. Though successful in compressing model, \cite{He2017, Zhao2019} originated in experience or intuition, lacking of basic theoretical guidance. \cite{Li2017} calculated the L1 norm of the filter, cutting out the feature map corresponding to the smaller L1 norm, and retrained after pruning. However, all layers use the same trim scale, which brings suboptimization. \cite{Lin2020} proposed an effective and efficient filter pruning approach that explored the High Rank (HRank) of the feature maps in each layer. The principle behind HRank is that low-rank feature maps contain less information, and thus pruned results can be easily reproduced. Nevertheless, similar high-rank feature maps may also contain redundant information, but they have not been removed. 

	Different from the existing methods, our method builds a bridge between tensor compression and model pruning, and provides a new perspective for filter pruning by deleting redundant information of tensor. With a new theoretical foundation, QSFM can get rid of the limitation of special software library or hardware, achieving the promotion of storage usage, memory occupation and computing speed at the same time. And it retains the model structure with different pruning rate. It can also be combined with other typical categorise mentioned above to further achieve higher compression ratio.

\begin{figure*}[t]
\begin{center}
   \includegraphics[width=0.8\linewidth]{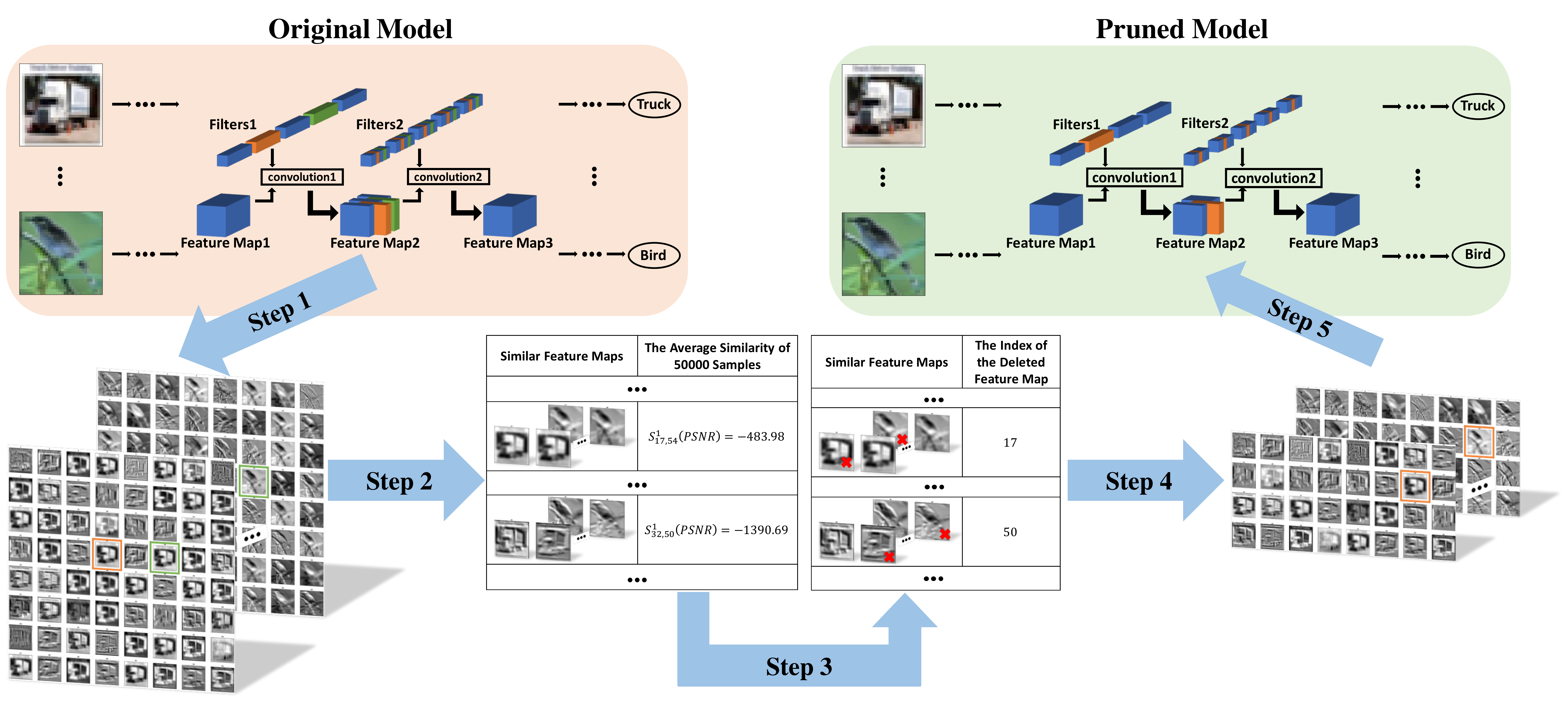}
\end{center}
\captionsetup{font={footnotesize}} 
\vspace{-0.3cm}
   \caption{This figure describes the detailed pruning workflow. Step 1, obtain feature maps generated by convolution layer. Step 2, quantify similarity between feature maps. Step 3, use auxiliary condition to decide which feature map should be deleted. Step 4, identify feature maps to keep. Step 5, delete filters.}
\vspace{-0.4cm}
\label{fig:X}
\end{figure*}

\section{Quantified Similarity between Feature Maps} \label{section3}

	In mathematics, a set of vectors is linearly independent if no vector in the set can be expressed as a linear combination of the other vectors. If $A$ is a linearly independent set of vectors and $B$ is linearly dependent, where $A$ and $B$ are composed of the same number and same dimensions of vectors, then there are more vectors that can be represented by $A$ than by $B$, and in this sense, $B$ contains redundant information.

	We extend this linearly independent concept to the three dimensional tensors composed of feature maps in CNNs. To some extent, every vector in a linearly independent set is not similar. For each convolution layer in CNNs, its filters are convolved with the input to generate feature maps. These 2D feature maps form 3D tensors and are then input to the next convolution layer. Our pipeline (Quantified Similarity between Feature Maps, QSFM) tries to identify similar feature maps to find out the redundant information of 3D tensors, and prunes them by deleting similar feature maps. The whole process is like constructing the maximum linearly independent group of 3D tensors. The detailed process is described as follows:

\subsection{Assumptions}

    The pre-trained CNN model before pruning is $Model_0$. Filter pruning does not change the convolution layers number, so we assume that the convolution layers of the model are $L_1, L_2, ..., L_n$. QSFM prune these $n$ layers sequentially, and the model obtained after the $k$th pruning operation is $Model_k$.

    The parameters set $W_i$ of the $i$th convolution layer $L_i$ is composed of $N_i$ filters, where filters are 3D tensors and $W_i$ is a 4D tensor. Denoted as  $W_i = \{F_{(i, 1)}, F_{(i, 2)}, ..., F_{(i, N_i)}\} \in \mathbb{R}^{N_i \times N_{i-1} \times K_i \times K_i}$, where $F_{(i, j)} \in \mathbb{R}^{N_{i-1} \times K_i \times K_i}$ represents the $j$th filters, $ N_{i-1}$ represents the channel number for a filter in $L_i$ (also the input channel number), $N_i$ represents the  output channel number for $L_i$ and $K_i$ represents the height and width of the convolution filter. Note that the output channel number of the convolution layer is equal to the number of filters contained in this layer, while the filter channel number in this layer is equal to the number of output channels of the previous layer, so the notation of $N_i$ and $N_{i-1}$ above is rigorous.

    Assume an image dataset $Train$ has $M$ images, denoted as $Train = \{Image_1, Image_2, ..., Image_k\} \in \mathbb{R}^{M \times N_0 \times X_0 \times Y_0}$, where $N_0$ represents the channel number, $X_0$, $Y_0$ are the width and height. If input only one image into the model, then the input of $L_i$ is $I_i \in \mathbb{R}^{N_{i-1} \times X_{i-1} \times Y_{i-1}}$ and the output of $L_i$ is $O_i \in \mathbb{R}^{N_i \times X_i \times Y_i}$. Actually $I_i = O_{i-1}$, but separating them out makes readers focus more on the input and output of a particular convolution layer ($L_i$). For a specific convolution operation in $L_i$, the $j$th filter $F_{(i, j)}$ in $W_i$ is convolved with $I_i$ to generate a 2D feature map, denoted as $F\_M_{(i,j)} \in \mathbb{R}^{X_i \times Y_i}$. All these $N_i$ filters in layer $L_i$ generate $N_i$ feature maps, denoted as $O_i = \{F\_M_{(i,1)},~F\_M_{(i,2)},~...$, $F\_M_{(i,N_i)}\} \in \mathbb{R}^{N_i \times X_i \times Y_i}$. 

    Above notations can refer to Fig.~\ref{fig2}, the yellow background represents the input or output of each layer and the blue background is the parameters of the model. As shown in Fig.~\ref{wzd1} and Fig.~\ref{fig2}, the detailed pruning operation has been clear, so the upper part of Fig.~\ref{fig:X} becomes more abbreviated, while the lower part focuses on the operations of QSFM on feature maps.

    It can be concluded that specific feature maps correspond to specific filters. QSFM inputs $M$ images in  $Train$  into the model to find redundant feature maps and delete filters using the corresponding relationship. Specifically, if QSFM is going to prune the $k$th convolution layer $L_k$ of $Model_{k-1}$, after identifying the redundant feature maps, QSFM delete the corresponding filters and obtain $Model_{k}$.

\subsection{Find Redundant Feature Maps}

    Filter pruning reduces the size and computation of CNN model by reducing the number of channels output by the convolutional layer. For the convolution layer $L_i$, that means the $N_i$ need to be reduced and the 3D tensor $O_i$ composed of feature maps need to be compressed correspondingly. Therefore, it is important to reduce the number of 3D tensor channels while preserving as much information as possible.

    Assume a compact 3D tensor  $O^{compact} = \{F\_M_{1}^{c},~F\_M_{2}^{c},~...$, $F\_M_{N_{c}}^{c}\} \in \mathbb{R}^{N_{c} \times X \times Y}$, where $F\_M_{j}^{c} \in \mathbb{R}^{X \times Y}$ is not similar to each other. $N_{r}$ elements in $O^{compact}$ are allowed to be repeatedly selected to form a redundant 3D tensor  $O^{redundant} \in \mathbb{R}^{N_{r} \times X \times Y}$. In general, an ordinary 3D tensor $O \in \mathbb{R}^{N \times X \times Y}$ is a combination of compact tensor $O^{compact}$ and redundant tensor $O^{redundant}$, where $N=N_{c}+N_{r}$. 

    If we extend the terms of the above $O^{redundant}$ composition process, allow $O^{redundant}$ to contain elements similar to those in $O^{compact}$ (not exactly equal), by visualizing the feature maps output by CNN convolutional layer, we can find that these 3D tensor output can be composed of a compact tensor and a redundant tensor (as shown in Fig.~\ref{wzd1} and  Fig.~\ref{fig:X}, similar results can be seen in GhostNet \cite{Han2020}). Since the elements in $O^{redundant}$ are approximately equal to the elements in $O^{compact}$ (similar), $O^{redundant}$ represents redundant information in $O$.

    QSFM uses the above properties to find similar feature maps and delete redundant feature maps to build compact tensor. This process is like constructing the maximum linearly independent group. QSFM does not care about how to measure the similarity between feature maps. Instead, it wants to compress 3D tensors by taking advantage of the similarity between feature maps. In order to show the operation process concretely, we select two similarity functions to measure similarity respectively. (The Step 2 in Fig.~\ref{fig:X} shows only one of these methods.)  However, readers can also customize the similarity function and even use neural network to judge whether the feature maps are similar.

\subsection{Quantify Similarity by Similarity Functions}

	The sequence of pruning is from $L_1$ to $L_n$, and it is assumed that the $i$th convolution layer $L_i$ is being pruned.

	First, assume the number of filters to be pruned in layer $L_i$ is $N_{i2}$ according to the compression rate (set manually according to requirements). These $N_{i2}$ filters which should be pruned constitute the set $Delete^i$. 

    In practice, QSFM makes quantified similarity more robust by averaging the results of multiple inputs rather than a single image.

	The model to be pruned is $Model_{i-1}$, when the $Image_k$ is input into the $Model_{i-1}$, the output of $L_i$ is $O_i^k = \{F\_M_{(i,1)}^k, F\_M_{(i,2)}^k, ...$, $F\_M_{(i,N_i)}^k\} $. Calculate the quantified similarity function $S(F\_M_{(i,m)}^k, F\_M_{(i,n)}^k)$ where $F\_M_{(i,m)}^k$ and $F\_M_{(i,m)}^k$ ($m\neq n$) are any two elements in the set $O_i^k$.  Hereinafter, referred to as $S_{m, n}^{i, k}$, function $S$ takes Structural Similarity (SSIM) \cite{Wang2004} or Peak Signal to Noise Ratio (PSNR). 

	SSIM and PSNR are usually used to measure the image quality after compression. As far as we know, this is the first time that these methods are used to measure the similarity between two feature maps in CNNs.

\begin{small}
\begin{equation}
\begin{aligned}
\begin{array}{l}
S_{m,n}^{i,k}(SSIM)\\
\\ 
{\kern 1pt} {\kern 1pt} {\kern 1pt} {\kern 1pt} {\kern 1pt} {\kern 1pt} {\kern 1pt} {\kern 1pt} {\kern 1pt} {\kern 1pt} {\kern 1pt} {\kern 1pt} {\kern 1pt} {\kern 1pt} {\kern 1pt} {\kern 1pt} {\kern 1pt} {\kern 1pt} {\kern 1pt}  \vspace{1ex}= SSIM(F\_M_{(i,m)}^k,F\_M_{(i,n)}^k)\\
{\kern 1pt} {\kern 1pt} {\kern 1pt} {\kern 1pt} {\kern 1pt} {\kern 1pt} {\kern 1pt} {\kern 1pt} {\kern 1pt} {\kern 1pt} {\kern 1pt} {\kern 1pt} {\kern 1pt} {\kern 1pt} {\kern 1pt} {\kern 1pt} {\kern 1pt} {\kern 1pt} {\kern 1pt}  = \dfrac{{\displaystyle (2{\mu _m}{\mu _n} + k_1^2{D^2})(2{\sigma _{mn}} + k_2^2{D^2})}}{{\displaystyle (\mu _m^2 + \mu _n^2 + k_1^2{D^2})(\sigma _m^2 + \sigma _n^2 + k_2^2{D^2})}}.
\end{array}
\end{aligned}
\end{equation}
\end{small}

	Here $\mu_m$ and $\mu_n$ are the average of all pixels in $F\_M_{(i,m)}^k$ and $F\_M_{(i,n)}^k$. $\sigma _m^2$ and $\sigma _n^2$ are the variance of all pixels in $F\_M_{(i,m)}^k$ and $F\_M_{(i,n)}^k$. $\sigma _{mn}$ is the covariance of $F\_M_{(i,m)}^k$ and $F\_M_{(i,n)}^k$. Usually $k_1$ is 0.01 and $k_2$ is 0.03 according to \cite{Wang2004}. $D$ is determined by the following equation:

\begin{small}
\begin{equation}
\begin{aligned}
D = \max (O_i^k) - \min (O_i^k).
\end{aligned}
\end{equation}
\end{small}
\\[-5ex]

	Here $\max (O_i^k)$ and $\min (O_i^k)$ are the values of the largest and smallest pixels in $(O_i^k)$.

	Since the quantization of PSNR is theoretically equivalent to the Euclidean distance if only care about the  value of similarity ordering, QSFM only need to sort the similarity according to the Euclidean distance, needless to know the specific value of PSNR. Therefore, the following content does not distinguish the PSNR from the Euclidean distance. In order to make the judgment standard of quantification function more uniform, we hope that the larger the value of the similarity function is, the more similar the two feature maps are. So we choose to take the negative of the Euclidean distance.

\begin{small}
\begin{equation}
\begin{aligned}
\begin{array}{l}
S_{m,n}^{i,k}(PSNR)\\
\\ 
{\kern 1pt} {\kern 1pt} {\kern 1pt} {\kern 1pt} {\kern 1pt} {\kern 1pt} {\kern 1pt} {\kern 1pt} {\kern 1pt} {\kern 1pt} {\kern 1pt} {\kern 1pt} {\kern 1pt} {\kern 1pt} {\kern 1pt} {\kern 1pt} {\kern 1pt} {\kern 1pt}  \vspace{1ex}= -EuclideanDistance(F\_M_{(i,m)}^k,F\_M_{(i,n)}^k)\\
{\kern 1pt} {\kern 1pt} {\kern 1pt} {\kern 1pt} {\kern 1pt} {\kern 1pt} {\kern 1pt} {\kern 1pt} {\kern 1pt} {\kern 1pt} {\kern 1pt} {\kern 1pt} {\kern 1pt} {\kern 1pt} {\kern 1pt} {\kern 1pt} {\kern 1pt} {\kern 1pt}  = -\sqrt{\sum\limits_{x = 1}^{{X_i}} {\sum\limits_{y = 1}^{{Y_i}} {{{(F\_M_{(i,m)}^k[x,y] - F\_M_{(i,n)}^k[x,y])}^2}} }}.
\end{array}
\end{aligned}
\end{equation}
\end{small}

    It is noted that $S_{m, n}^{i, k}$ is only the result generated by a single Image ($Image_k$). For the whole $Train$, it contains $M$ images in total. We calculate the statistical average results on $M$ images, and define: 
\\[-1ex]
\begin{small}
\begin{equation}
\begin{aligned}
\begin{array}{l}
S_{m,n}^i = \dfrac{{\displaystyle \sum\limits_{k = 1}^M {S_{m,n}^{i,k}} }}{\displaystyle M},\\ \\ {S^i} = \{ S_{m,n}^i\} ,(m,n = 1,2,3,...,{N_i}).
\end{array}
\end{aligned}
\end{equation}
\end{small}
\subsection{Prune Convolution Layers by Deleting Filters}
	After finding out the feature map groups with high similarity, we need to stipulate an auxiliary condition to determine which one of the two feature maps to delete. (Step 3 in Fig.~\ref{fig:X}) We can randomly delete a feature map in each similar group, or we can use the L1 norm of the feature map to delete those with a smaller norm. We use the rank of the two-dimensional matrix as auxiliary condition to assist the quantified similarity function finding the redundant feature maps. The use of this auxiliary condition is inspired by HRank \cite{Lin2020}, but it does not mean that QSFM is similar to that of HRank. In fact, as mentioned above, HRank does not take into account the existence of redundant information (i.e. similarity) in two high-rank feature maps. Meanwhile, our experiment will prove that QSFM is superior to HRank. Calculate the rank of each $F\_M_{(i,m)}^k$, then  the statistical average of the rank of the $F\_M_{(i,m)}^k$ in $O_i$ is:

\begin{small}
\begin{equation}
\begin{aligned}
\begin{array}{l}
Rank_m^i = \dfrac{{\displaystyle\sum\limits_{k = 1}^M {Rank_m^{i,k}} }}{\displaystyle M},\\
\\  Ran{k^i} = \{ Rank_m^i\} ,(m = 1,2,3,...,{N_i})
\end{array}
\end{aligned}
\end{equation}
\end{small}
	
\begin{figure}[t]
\begin{center}
   \includegraphics[width=0.8\linewidth]{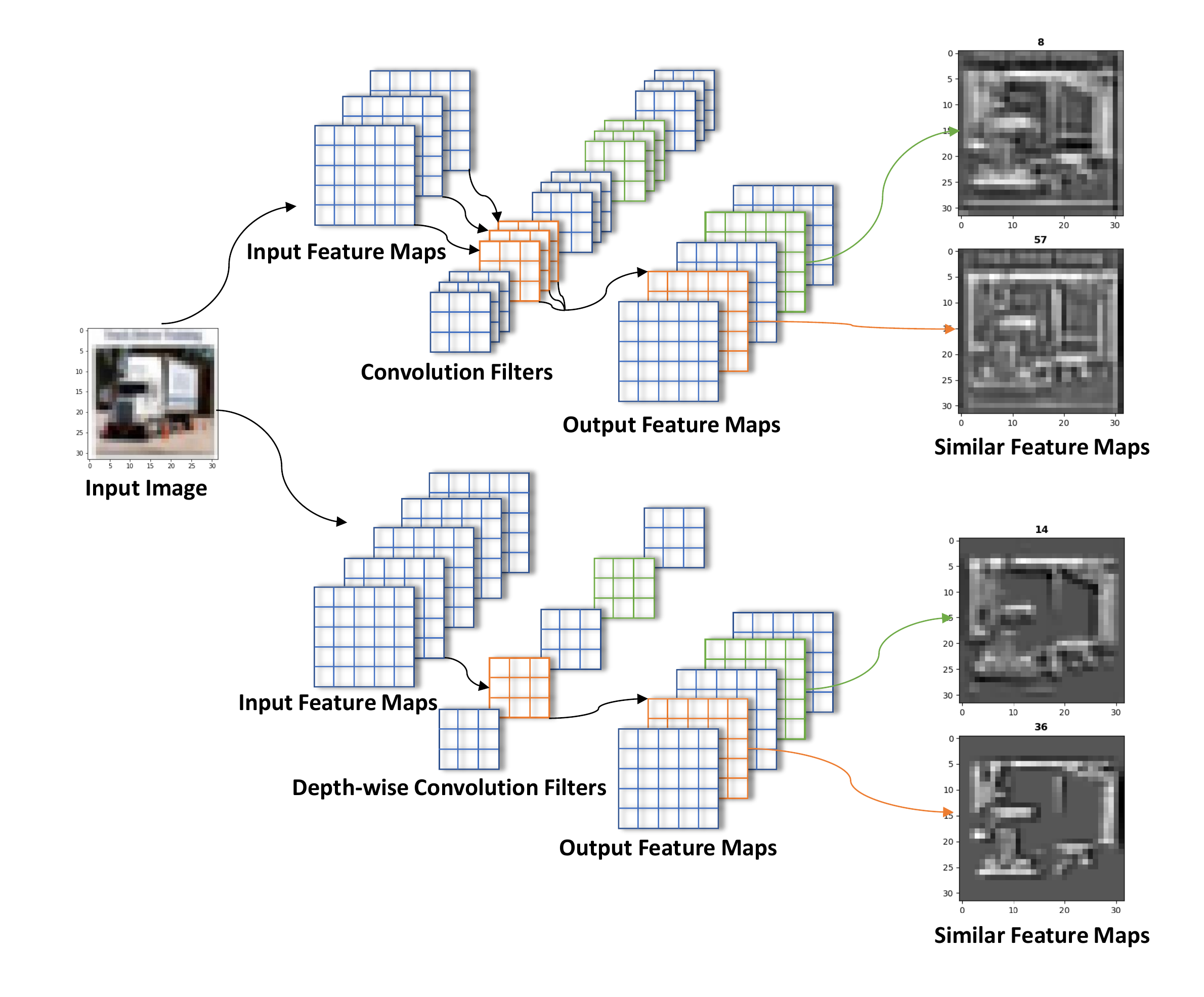}
\end{center}
\captionsetup{font={footnotesize}} 
\vspace{-0.4cm}
   \caption{The specific operation of common convolution and depth-wise separable convolution is slightly different, but QSFM can prune both of them.}
\vspace{-0.5cm}
\label{fig:2}
\end{figure}

	Arrange the set $S^i$ from high to low, then match the set $Rank^i$ to determine the filter to be pruned. Set the conditions as follows:
\begin{itemize}
	\item Condition 1: $Filter_{(i,m)}$ and $Filter_{(i,n)}$ are not members of $Delete^i$;
	
	\item Condition 2: $S_{m,n}^i$ is maximum under the premise that $m$ and $n$ meet condition 1;
	
	\item Condition 3: $Rank_m^i \textgreater Rank_n^i$;

	\item Condition 4: $Rank_m^i \leq Rank_n^i$;
\end{itemize}

	If conditions 1, 2 and 3 meet simultaneously, $Filter_{(i,n)}$ will be put into $Delete^i$; if conditions 1, 2 and 4 meet simultaneously, $Filter_{(i,m)}$ will be put into $Delete^i$. And the above operation will continue until the number of filters in the set $Delete^i$ is $N_{i2}$. (Step 4 and Step 5 in Fig.~\ref{fig:X}) 

	The model could be fine-tuned after pruning every convolution layer, that is:
\begin{small}
\begin{equation}
\begin{aligned}
Model_i = Fine\_Tuned(Mode{l_{i - 1}} - Delet{e^i})
\end{aligned}
\end{equation}
\end{small}
\\[-5ex]

	After all convolutional layers are pruned following the above methond, the compression of the neural network is complete successfully.

    The rank is not the only auxiliary condition can be used. In fact, if $S_{m,n}^i$ is large, it indicates that there is similar redundant information between the $m$th feature map and the $n$th feature map, and the auxiliary condition is to determine whether to delete the $m$th or the $n$th feature map. Section III.C. is to quantify similar redundant information between feature maps, while Section III.D. is to determine how to delete redundant information (especially for neural network pruning). As long as a method can play the same role, it can be used as the auxiliary condition of QSFM.

\subsection{Special Convolution}

\begin{figure}[t]
\begin{center}
   \includegraphics[width=0.9\linewidth]{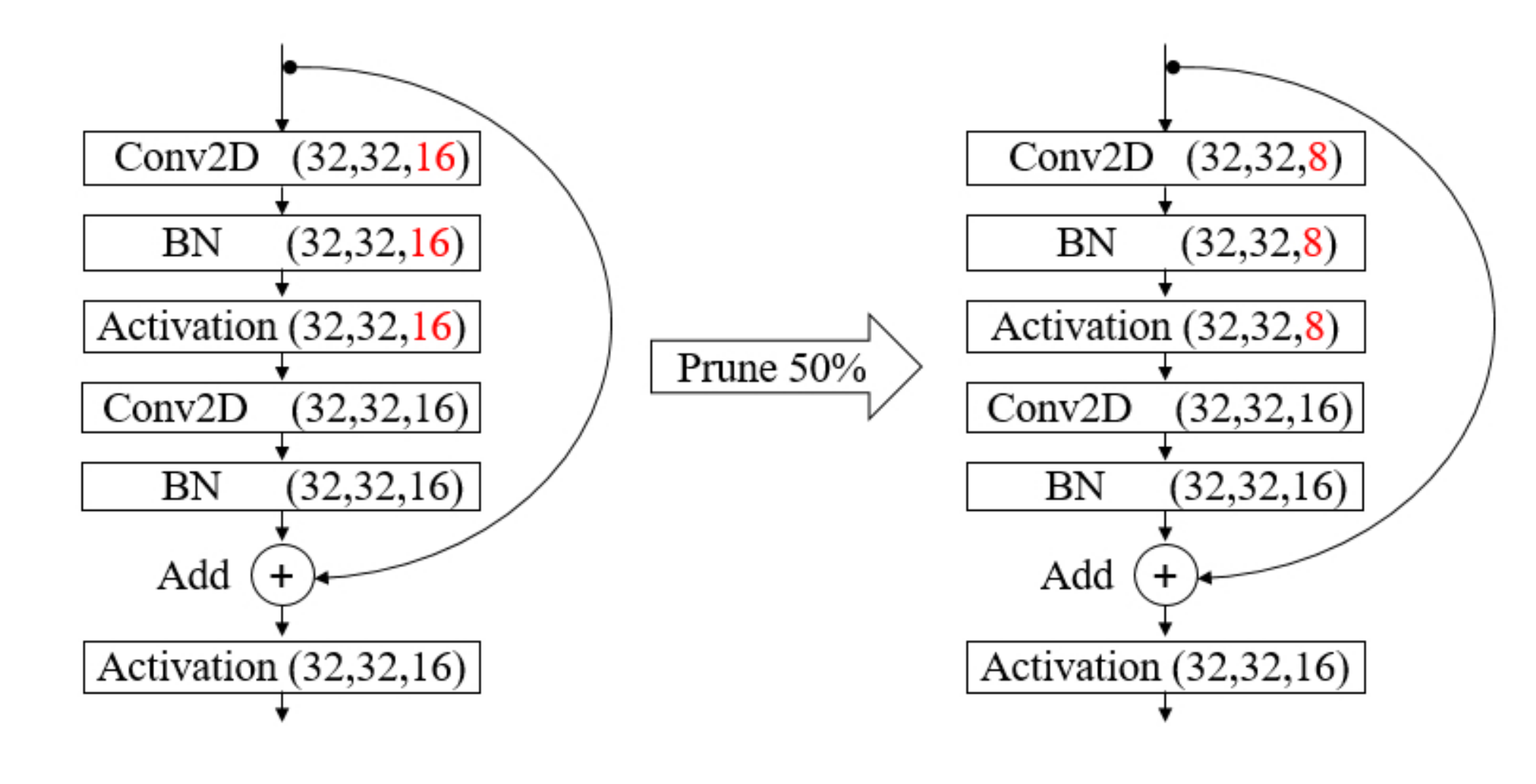}
\end{center}
\captionsetup{font={footnotesize}} 
\vspace{-0.5cm} 
   \caption{The pruning strategy of ResNet-56.}
\vspace{-0.3cm} 
\label{fig:3}
\end{figure}
\begin{figure}[t]
\begin{center}
   \includegraphics[width=0.9\linewidth]{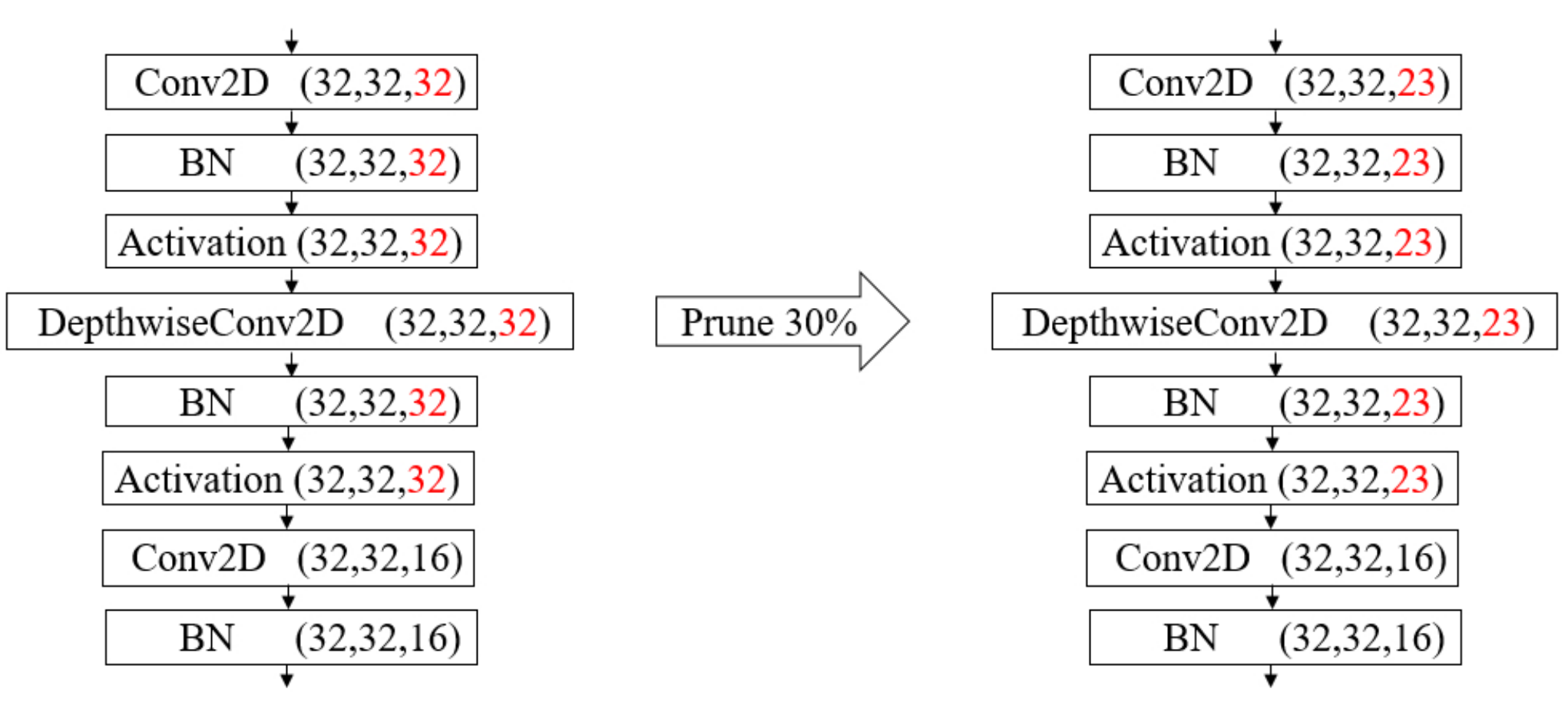}
\end{center}
\captionsetup{font={footnotesize}} 
\vspace{-0.45cm}
   \caption{The pruning strategy of MobileNet-V2.}
\vspace{-0.3cm} 
\label{fig:4}
\end{figure}

	QSFM can also prune special convolution layers, like depth-wise separable convolution in MobileNet. As shown in Fig.~\ref{fig:2}, no matter how the convolution layer is convolved, the corresponding relationship between filters and feature maps remains unchanged and the output still have similar feature maps. So QSFM can prune these special CNNs.

\section{Experiments}

    In this section, we give some experimental settings firstly and prove the feasibility of QSFM through ablation study. Then we use use different model structures to test QSFM's performance on different datasets. Finally, we deploy the pruned model of QSFM in edge devices, and verify that QSFM is helpful to AI on Edge tasks by accelerate CNNs inference speed.

	In our experiments, we use Euclidean distance and Structural Similarity as the function of quantifying the similarity for QSFM, which are called as QSFM-PSNR and QSFM-SSIM respectively. 

\subsection{Experimental Settings}

\subsubsection{Datasets}

CIFAR-10 \cite{CIFAR10} consists of 60000 color images with a resolution of 32 ${\times}$ 32. The images in CIFAR-10 are labeled as 10 categories, each containing 6000 images, 5000 for training and 1000 for testing. CIFAR-100 \cite{CIFAR10} consists of 60000 color images with a resolution of 32 ${\times}$ 32. The images in CIFAR-100 are labeled as 100 categories, each containing 600 images, 500 for training and 100 for testing.

Due to the appropriate amount of data and small resolution, CIFAR-10 and CIFAR-100 are widely used by many image classification algorithms. However, because of its low resolution, it is used to test the performance of algorithms in most cases, not in actual deployment scenarios.

ILSVRC-12 \cite{imagenet} contains more than 1.28 million images, which vary in resolution but are usually adjusted to 224 ${\times}$ 224 resolution for use. The images are labeled as 1000 categories.

Different from CIFAR-10 and CIFAR-100, ILSVRC-12 has a large enough amount of data, high resolution and multiple categories to ensure that it can be used in actual deployment scenarios.

\subsubsection{Model Structures}

    The model structures we used are VGG-16 \cite{Simonyan2015}, ResNet-56 \cite{He2016}, and MobileNet-V2 \cite{Sandler2018}. For CIFAR-10, the initial model (baseline) used by QSFM are VGG-16 (top-1 accuracy 93.39\%), ResNet-56 (top-1 accuracy 93.21\%) and  MobileNet-V2 (top-1 accuracy 92.54\%). For CIFAR-100, the baseline is ResNet-56 (top-1 accuracy 70.62\%). For ILSVRC-12, the baseline is MobileNet-V2 (top-1 accuracy 72.21\%).

\subsubsection{Configurations}
    Images in CIFAR-10 and CIFAR-100 have resolutions of 32 ${\times}$ 32 ${\times}$ 3, while images in ILSVRC-12 are resized to 224 ${\times}$ 224 ${\times}$ 3. 

	All the pruning operation are conducted within Tensorflow(1.14.0) and Keras(2.2.5). In practical application, we use TensorFlow Lite to apply the model on mobile devices.

	The convolution operation is usually followed by batch normalization layer and activation layer, in our experiments, we regard these three layers as a block and conduct QSFM pruning on its final output 3D tensor(feature maps).

	For every residual block in ResNet-56, we only prune the first convolutional layers, which are simple and can keep the output dimension of residual block unchanged, as shown in Fig.~\ref{fig:3}. For every  bottleneck in MobileNet-V2, we prune depth-wise separable convolution layers, as shown in Fig.~\ref{fig:4}. 

	After the whole pruning operation, we calculate the FLOPs and Parameters of the pruned models and compare them with existing methods \cite{Lin2020, He2018, He2017, Lin2019, Zhao2019, cuili}. On edge devices, we further measure the inference speed to measure QSFM's performance.

\subsubsection{Devices}
    We use a server with a Nvidia V100 GPU to compress CNNs by QSFM. For edge divice, we use a Xiaomi-M2006J10C with an ARM MT6889Z/CZA processor to deploy TensorFlow Lite model and a Nvidia-Jetson-TX2 with a 256-core Pascal GPU to deploy TensorFlow model. Note that even if on the same device, the inference speed of the same model may still be different in different software environments or physical environments, so the measurement of the inference speed of CNNs can be regarded as a qualitative experiment, while FLOPS and Parameters are more universal because they are not affected by equipment and environment.

\begin{figure}[t]
\begin{center}
   \includegraphics[width=1.0\linewidth]{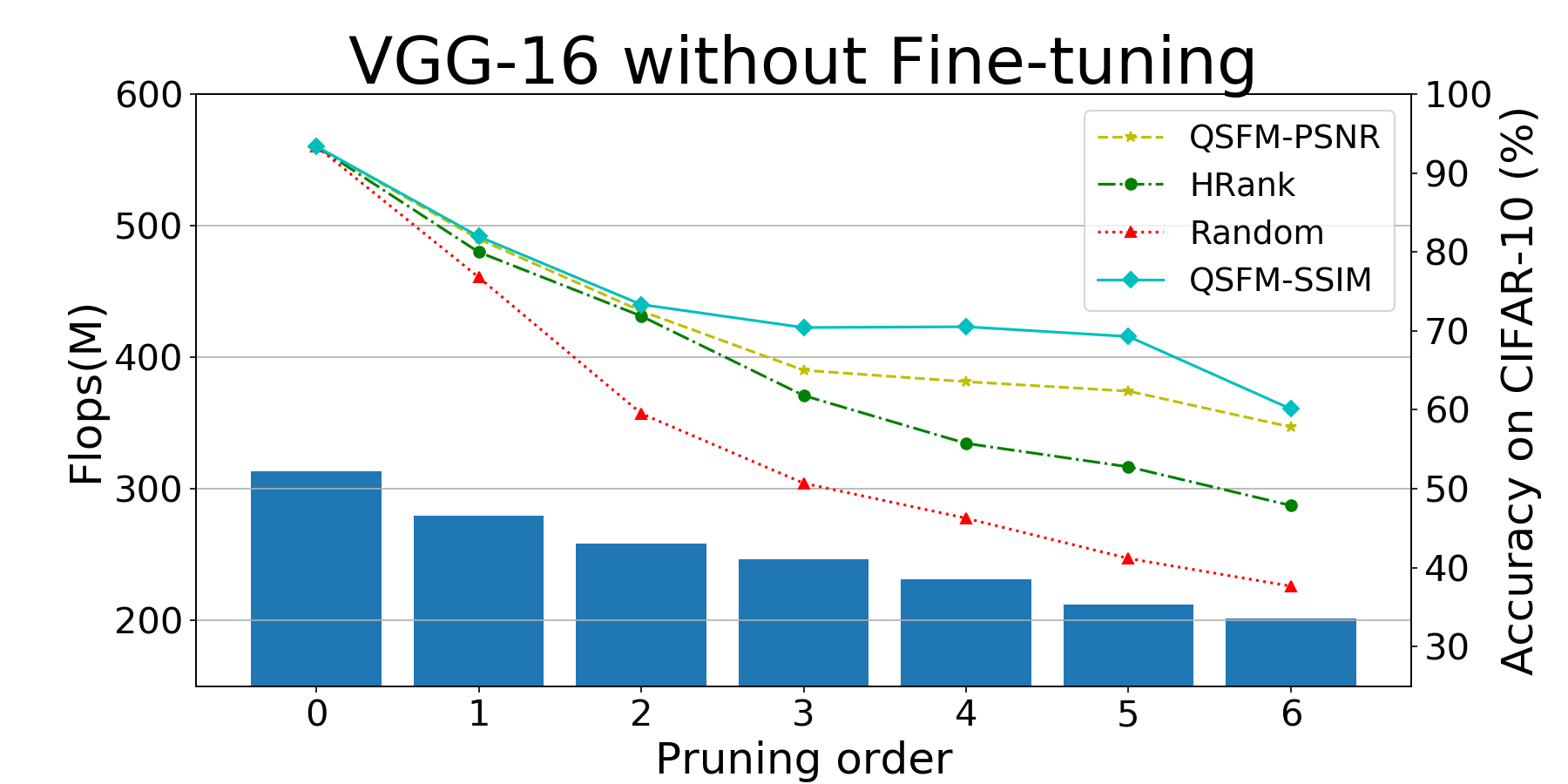}
\end{center}
\captionsetup{font={footnotesize}} 
\vspace{-0.3cm}
   \caption{The histogram shows the FLOPs of the model after each pruning. The curve shows the change of model accuracy after each pruning under four different methods.}
\vspace{-0.3cm}
\label{vgg-ablation}
\end{figure}

\subsection{Ablation Study}
    To verify that QSFM can find the redundant information more efficiently in feature maps, we prune VGG-16 and ResNet-56 without fine-tuning under the same compression rate and other conditions, and compared with random pruning and existing method (Hrank).
\subsubsection{VGG-16}
	For VGG-16 on CIFAR-10, QSFM prune the $3^{th}$ and $4^{th}$ blocks(the $5^{th}$ to $10^{th}$ convolution layers), and the compression rates are [0.6, 0.4, 0.3, 0.3, 0.3, 0.3]. At this compression ratio, we prune the network layer by layer with different methods, including QSFM-PSNR, QSFM-SSIM, Hrank, Random(delete filters randomly). It should be emphasized that we did not make any fine-tuning in this part, and the experimental results are shown in the Fig.~\ref{vgg-ablation}.

Our method is consistently better than Hrank and random method in pruning different convolutional layers. In the experiment, we find that sometimes the accuracy of our method after pruning is even higher than that before pruning(In steps $3^{th}$ and $4^{th}$ of QSFM-SSIM). When the compression ratio is large, HRank needs to delete the high-rank feature maps, but it cannot effectively distinguish which are redundant in these high-rank feature maps.

\begin{figure}[t]
\begin{center}
   \includegraphics[width=1.0\linewidth]{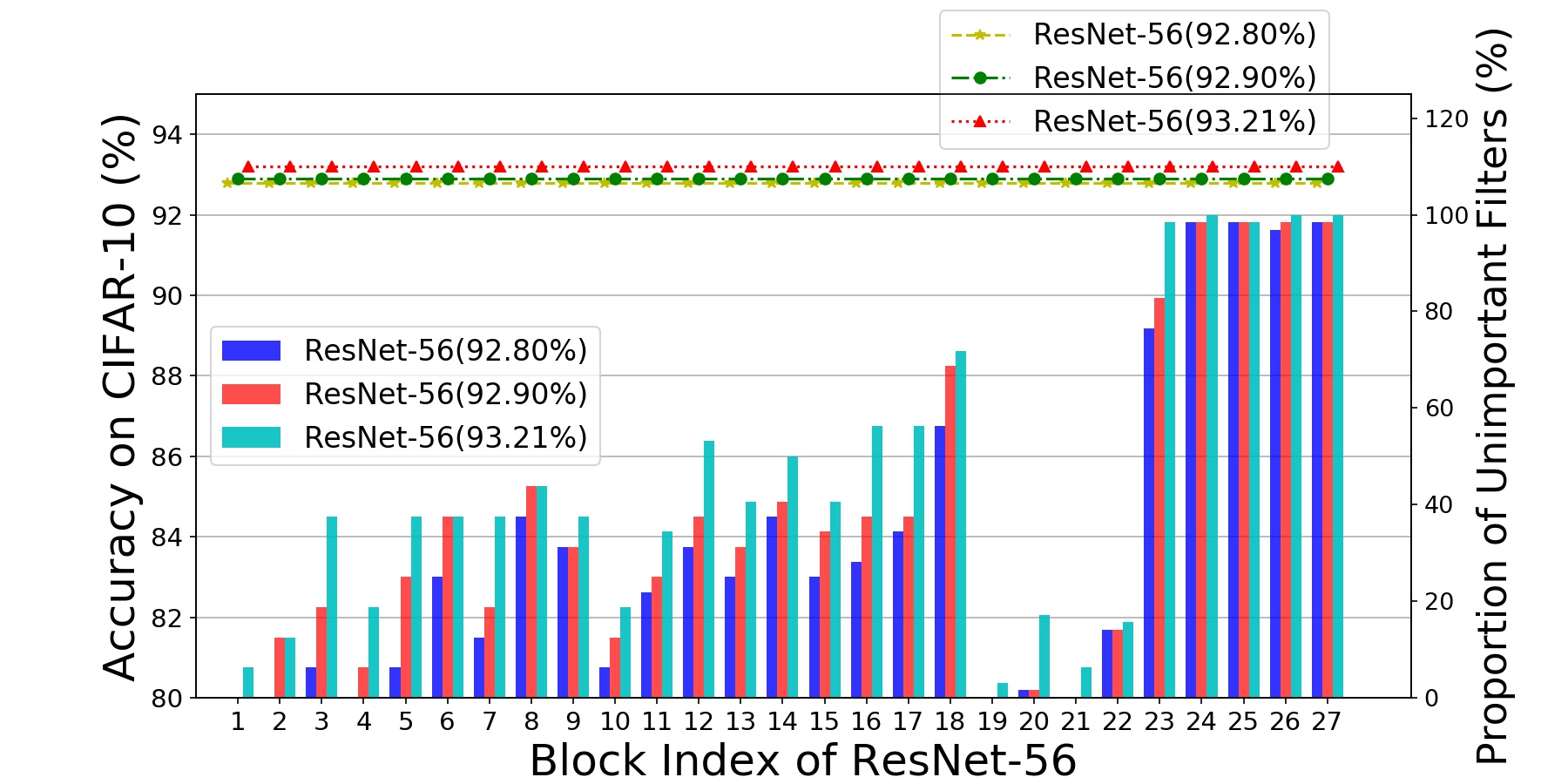}
\end{center}
\captionsetup{font={footnotesize}} 
\vspace{-0.2cm}
   \caption{The x-axis represents the block index of ResNet-56. The coordinates of histogram represent the proportion of ‘unimportant filters' in each layer of the three models. The curve represents the change of the accuracy of the model after each pruning, and it can be found that the accuracy of the model has not changed from beginning to end.}
\vspace{-0.3cm}
\label{resnet-ablation1}
\end{figure}

\begin{figure}[t]
\begin{center}
   \includegraphics[width=1.0\linewidth]{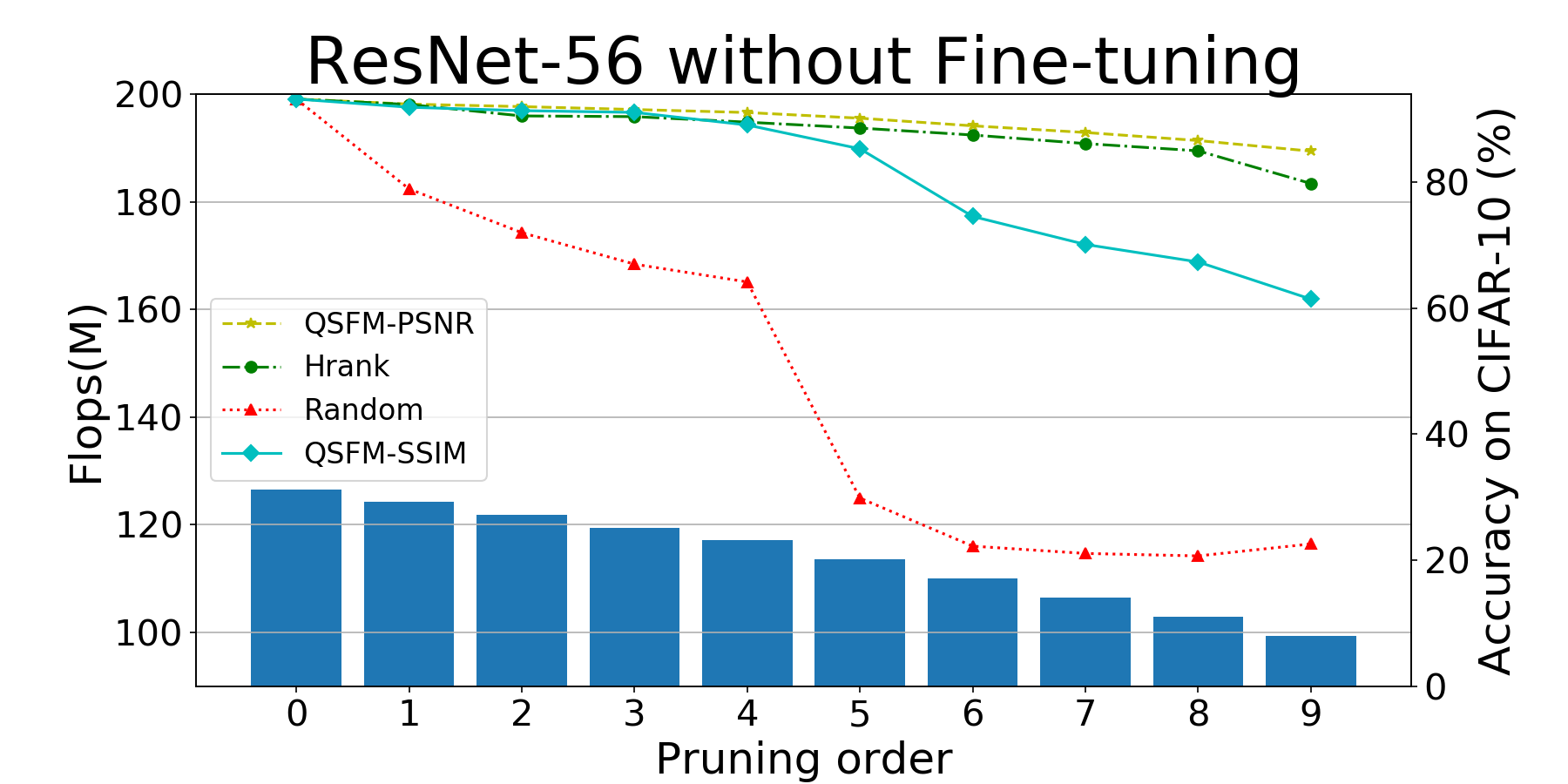}
\end{center}
\captionsetup{font={footnotesize}} 
\vspace{-0.2cm}
   \caption{The histogram shows the FLOPs of the ResNet-56 model after each pruning. The curve shows the change of model accuracy after each pruning under four different methods.}
\vspace{-0.3cm}
\label{resnet-ablation2}
\end{figure}

\subsubsection{ResNet-56}

    In ResNet-56 on CIFAR-10, we find quite a few convolution layers have many filters whose all parameter values close to $10^{-32}$ ('unimportant filters'). We count the proportion of filters whose all parameter values are close to  $10^{-32}$ in the 27 layers of every residual block and prune these layers. When we prune these filters, the accuracy of the pruned model was roughly the same as that of the original model (without fine-tuning). In order to prove the generality of the results, we trained ResNet-56 on CIFAR-10 by three independent training with different accuracy (92.80\%, 92.90\% and 93.21\%), all of which have similar results, as shown in Fig.~\ref{resnet-ablation1}. This shows that there is a lot of redundancy in the model of ResNet-56 on CIFAR-10, so we think that it may lead to the selection of filters that should be pruned are all of magnitude $10^{-32}$ if the compression ratio of each convolutional layer is too small, which cannot reflect the advantages and disadvantages of our method.

   To sum up, we can only judge the performance of the method if the compression ratio is high enough (larger than the proportion of filters whose all parameter values close to $10^{-32}$). If the compression ratio is too low, then any method will always get a good result because the filters they delete are distinctly useless.

    We use this 'high compression ratio' principle to compare the pruning performance of various methods (the compression ratio of each layer of ResNet-56 on CIFAR-10 will not be lower than 0.5).

     We prune the first 9 blocks of ResNet-56, and the compression rates are [0.5, 0.5, 0.5, 0.5, 0.75, 0.75, 0.75, 0.75, 0.75]. We use the whole training set (50000 images) to calculate the similarity between feature maps. The results are shown in Fig.~\ref{resnet-ablation2}. Compared with Hrank and Random, QSFM-PSNR provides better maintenance of precision for the whole 9 steps, which demonstrates QSFM can better identify important filters. But for QSFM-SSIM, it only performed better in steps $1^{th}$ to $3^{th}$ , which indicates that there is still space for improvement in the selected function to measure similarity. Both QSFM-PSNR and QSFM-SSIM are far better than Random, which proves the correctness of our method.

\begin{table}
\scriptsize
\captionsetup{font={footnotesize}}
\caption{Pruning results of VGG-16 on CIFAR-10 (with Fine-tuning). 'Top-x' represents the Top-x accuracy. 'PR' represents the pruning rate. The accuracy is represented in the format of Baseline ${\rightarrow}$ Pruned (Decreased Accuracy). We use boldface to denote our method. The other tables and figures follows the same convention.}
\begin{center}
\addvbuffer[-5pt -13pt]{
\setlength{\tabcolsep}{2mm}{
\begin{tabular}{c|c|c|c}
\toprule
Method & Top-1(\%) & FLOPs(PR) &  Parameters(PR) \\
\hline\midrule
EasiEdge-82\% \cite{cuili} & 93.73${\rightarrow}$93.42(-0.31) & 70.42M(77.6\%) & 0.58M (96.1\%) \\
HRank \cite{Lin2020} & 93.96${\rightarrow}$91.23(-2.73) & 73.67M(76.5\%) & 1.75M (88.1\%) \\
\textbf{QSFM-SSIM} & 93.39${\rightarrow}$92.17(-1.22) & 79.0M(74.8\%) & 3.68M(75.0\%) \\
\textbf{QSFM-PSNR} & 93.39${\rightarrow}$92.00(-1.39)  & 42.46M(86.5\%) & 0.49M(96.7\%)\\

\bottomrule
\end{tabular}}}
\end{center}
\label{table-1}
\end{table}

\subsection{Results and Analysis on CIFAR-10}

    We prune some mainstream models on CIFAR-10, including VGG-16, ResNet-56 and MobileNet-V2.

\subsubsection{VGG-16}
	We apply QSFM to prune the VGG-16 model with Fine-tuning. For QSFM-SSIM, all 13 convolution layers have a compression ratio of 0.5. For QSFM-PSNR, the 13 convolution layers' compression ratios are [0.5 0.5 0.5 0.5 0.5 0.6 0.6 0.6 0.9 0.9 0.9 0.9]. The results are displayed in Tab.~\ref{table-1}. Compared with HRank, QSFM-SSIM nearly compress same FLOPs while maintaining a better accuracy drop ( -1.22\% vs. -2.73\% ).  Compared with EasiEdge-82\% , QSFM-PSNR has larger FLOPs reduction(42.46M vs. 70.42M).

\begin{table}
\scriptsize
\captionsetup{font={footnotesize}} 
\caption{Pruning results of ResNet-56 on CIFAR-10(without Fine-tuning)}
\begin{center}
\addvbuffer[-5pt -13pt]{
\setlength{\tabcolsep}{1mm}{
\begin{tabular}{c|c|c|c}
\toprule
Method & Top-1(\%) & FLOPs(PR) &  Parameters(PR) \\
\hline\midrule
AMC \cite{He2018} & 92.80${\rightarrow}$90.10(-2.70) & 63.28M(50.0\%) & – \\
He \cite{He2017} & 92.80${\rightarrow}$90.80(-2.00)  & 62.52M(50.6\%) & –\\
GAL-0.6 \cite{Lin2019} & 93.26${\rightarrow}$92.98(-0.28)  & 78.30M(37.6\%) & 0.75M(11.80\%)\\
GAL-0.8 \cite{Lin2019} & 93.26${\rightarrow}$90.36(-2.90)  & 49.99M(60.2\%) & 0.29M(65.90\%)\\
Zhao \cite{Zhao2019} & 93.04${\rightarrow}$92.26(-0.78)  & 100.86M(20.3\%) & 0.68M(20.49\%)\\ 
\textbf{QSFM-SSIM} & 93.21${\rightarrow}$92.67(-0.54) & 64.92M(48.7\%) & 0.36M(57.9\%) \\
\bottomrule
\end{tabular}}}
\end{center}
\label{table-2}
\end{table}

\begin{table}
\scriptsize
\captionsetup{font={footnotesize}} 
\vspace{-0.3cm}
\caption{Pruning results of ResNet-56 on CIFAR-10(with Fine-tuning).}
\begin{center}
\addvbuffer[-5pt -13pt]{
\setlength{\tabcolsep}{1mm}{
\begin{tabular}{c|c|c|c}
\toprule
Method & Top-1(\%) & FLOPs(PR) &  Parameters(PR) \\
\hline\midrule
GAL-0.8 \cite{Lin2019} & 93.26${\rightarrow}$91.58(-1.68) &50.37M(60.2\%) & 0.29M(65.9\%)\\
GAL-0.6 \cite{Lin2019} & 93.26${\rightarrow}$93.38(0.12) & 78.30M(37.6\%) & 0.75M(11.8\%)\\
AMC \cite{He2018} & 92.80${\rightarrow}$91.90(-0.90) & 63.28M(50.0\%) & –\\
Hrank \cite{Lin2020} & 93.26${\rightarrow}$90.72(-2.54) & 32.77M(74.1\%) & 0.27M(68.1\%)\\
EasiEdge-30\% \cite{cuili} & 93.92${\rightarrow}$93.61(-0.31) & 56.93M(55.0\%) & 0.45M (47.1\%) \\
\textbf{QSFM-SSIM-1} & 93.21${\rightarrow}$91.92(-1.29)	& 53.15M(58.0\%) & 0.26M(69.1\%)\\
\textbf{QSFM-SSIM-2} & 93.21${\rightarrow}$91.88(-1.33) & 50.62M(60.0\%) & 0.25M(71.3\%)\\
\textbf{QSFM-PSNR} & 93.21${\rightarrow}$91.98(-1.23) & 53.15M(58.0\%) & 0.26M(69.1\%)\\
\bottomrule
\end{tabular}}}
\end{center}
\label{table-3}
\end{table}

\begin{table}
\scriptsize
\captionsetup{font={footnotesize}} 
\caption{Pruning results of MobileNet-V2 on CIFAR-10(with Fine-tuning)}
\begin{center}
\addvbuffer[-5pt -13pt]{
\setlength{\tabcolsep}{2mm}{
\begin{tabular}{c|c|c|c}
\toprule
Method & Top-1(\%) & FLOPs(PR) &  Parameters(PR) \\
\hline\midrule
MobileNet-V2 & 92.54\% & 78.38M(0.0\%) & 2.20M(0.0\%)\\
\textbf{QSFM-SSIM} & 92.09\% & 57.27M(26.93\%) & 1.67M(24.09\%)\\
\textbf{QSFM-PSNR} & 92.06\% & 57.27M(26.93\%) & 1.67M(24.09\%)\\
\bottomrule
\end{tabular}}}
\end{center}
\label{table-4}
\end{table}

\subsubsection{ResNet-56}
    First,we prune ResNet-56 without any fine-tuning operations according to the proportion which is slightly higher than that of the unimportant filters(filters whose all parameter values close to $10^{-32}$), and finally get good results as shown in Tab.~\ref{table-2}. Compared with AMC and He te al., QSFM-SSIM nearly compresses same FLOPs while maintaining a less accuracy drop (-0.54\% vs. -2.7\% and -0.54\% vs. -2.0\%). Compared with GAL-0.6, though QSFM-SSIM has a little higher accuracy drop(-0.54\% vs. -0.28\%), but it gains a larger FLOPs and parameters reduction(64.92M vs. 78.30M and 0.36M vs. 0.75M). Compared with GAL-0.8, QSFM-SSIM gains a less accuracy drop(-0.54\% vs. -2.9\%) while has a little shortage at compression aspect. Compared with Zhao et al., QSFM-SSIM has a less accuracy drop(-0.54\% vs. -0.78\%), larger FLOPs reduction(64.92M vs. 100.86M) and larger parameters reduction(64.92M vs. 78.30M and 0.36M vs. 0.68M).

    Then we further apply our method to prune the ResNet-56 model with Fine-tuning for higher compression. The results are shown in Tab.~\ref{table-3}. There is a slight difference in compression between QSFM-SSIM-1 and QSFM-SSIM-2. QSFM-PSNR again demonstrates its ability to obtain a high accuracy of 91.98\%, with 69.1\% parameters reduction and 58.0\% FLOPs reduction. This is significantly better than GAL-0.8. Compared with AMC, which obtains 91.90\% of top-1 accuracy and 50\% FLOPs reduction, QSFM-PSNR gets a larger FLOPs reduction(53.15M vs. 63.28M). Compared with Hrank, QSFM-SSIM-2 gains a less top-1 accuracy drop and parameters reduction( -1.33\% vs. -2.54\% and 0.25M vs. 0.27M). Compared with EasiEdge-30\% , all the three QSFMs have a larger FLOPs and parameters reduction with tolerable drop of accuracy.

\subsubsection{MobileNet-V2}

	For MobileNet-V2, we use QSFM to prune the whole 17 depthwise separable convolution layers on CIFAR-10. The compression ratio of each layer is 0.3. The results of pruning experiment are shown in Tab.~\ref{table-4}. After pruning, the accuracy of the model only dropped by 0.45\% (92.54\% $\rightarrow$ 92.09\%, QSFM-SSIM), with 26.93\% FLOPs and 24.09\% parameters reduction. The results also show that our proposed method can also achieve good results for depthwise separable convolution layers.

\begin{table}

\scriptsize
\captionsetup{font={footnotesize}} 
\caption{Pruning results of ResNet-56 on CIFAR-100(with Fine-tuning)}
\begin{center}
\addvbuffer[-5pt -13pt]{
\setlength{\tabcolsep}{1mm}{
\begin{tabular}{c|c|c|c|c}
\toprule
Method & Top-1(\%) & Top-5(\%) & FLOPs(PR) &  Parameters(PR) \\
\hline\midrule
ResNet-56 & 70.62\% & 92.00\% & 126.55M(0.0\%) & 0.86M(0.0\%)\\
\textbf{QSFM-SSIM} & 68.36\% & 90.90\% & 58.38M(53.87\%) & 0.42M(51.16\%)\\
\textbf{QSFM-PSNR} & 68.33\% & 91.08\% & 58.38M(53.87\%) & 0.42M(51.16\%)\\
\bottomrule
\end{tabular}}}
\end{center}
\label{table:5}
\end{table}

\subsection{Results and Analysis on CIFAR-100}

	QSFM only prunes the first convolutional layers for every residual block in ResNet-56 on CIFAR-100, and the final results are shown in Tab.~\ref{table:5}. The Top-1 accuracy of QSFM-SSIM dropped by 2.26\% (70.62\%$\rightarrow$68.36\%) and the Top-5 accuracy of QSFM-PSNR dropped by 0.92\% (92.00\%$\rightarrow$91.08\%), with 53.87\% FLOPs reduction and 51.16\% parameters reduction.

\subsection{Practical Application}

    For practical applications of IoT, we use ILSVRC-12 to train MobileNet-V2 and deploy it to edge devices for image classification task. MobileNet is a well-known lightweight CNN model used for many real-time computer vision tasks such as image classification, object detection and so on. We use QSFM to compress MobileNet-V2 to accelerate CNNs inference speed and verify that our method is helpful to AI on Edge tasks.

\begin{table}

\scriptsize
\captionsetup{font={footnotesize}} 
\caption{Pruning results of MobileNet-V2 on ILSVRC-12(with Fine-tuning)}
\begin{center}
\addvbuffer[-5pt -13pt]{
\setlength{\tabcolsep}{1mm}{
\begin{tabular}{c|c|c|c|c}
\toprule
Method & Top-1(\%) & Top-5(\%) & FLOPs(PR) &  Parameters(PR)\\
\hline\midrule
MobileNet-V2 & 72.21\% & 89.93\% & 3465.63M(0.0\%) & 1.79M(0.0\%) \\
\textbf{QSFM-PSNR} & 70.98\% & 88.39\% & 2088.70M(39.73\%) & 1.07M(39.89\%)\\
\bottomrule
\end{tabular}}}
\end{center}
\label{table-6}
\end{table}

\begin{table}
\scriptsize
\captionsetup{font={footnotesize}} 
\caption{Latency of MobileNet-V2 on edge devices}
\begin{center}
\addvbuffer[-5pt -13pt]{
\setlength{\tabcolsep}{1mm}{
\begin{tabular}{c|c|c|c}
\toprule
Method & \thead{Latency(ms)\\ Xiaomi-M2006J10C} &\thead{Latency(ms)\\ TX2-GPU}  &\thead{Latency(ms)\\ TX2-CPU}\\
\hline\midrule
MobileNet-V2 & 42.27 & 22.97 & 63.23\\
\textbf{QSFM-PSNR} & 33.27(1.27$\times$) & 15.03(1.53$\times$) & 43.35(1.46$\times$)\\
\bottomrule
\end{tabular}}}
\end{center}
\label{table-7}
\end{table}

    Few works prune the MobileNet model. We only found some data from AMC\cite{He2018} that could be compared with QSFM. AMC can compress MobileNet-V2 with 30\%  FLOPs reduction and 1.0\% Top-1 accuracy drop, while more detailed performance of QSFM can refer to the Tab.~\ref{table-6}.
    For mobile device, we use the TensorFlow Lite Converter to convert a traditional TensorFlow model into a TensorFlow Lite model to deploy. For Nvidia-Jetson-TX2, we deploy it directly using the Tensorflow model. In  practical applications, the inference speed of the model is also affected by software version and physical environment, so this is only a qualitative experiment. In order to ensure the accuracy of the measurement latency as much as possible, we averaged the inference speed of 1000 images, as shown in the Tab.~\ref{table-7}. Even for an enough lightweight CNN model like MobileNet-V2, QSFM can compress it and accelerate inference speed in practical tasks.

\section{Conclusion and Outlook}

	In this paper, we propose a novel theory to find the redundant information in three-dimensional tensors, named QSFM. We apply QSFM to prune CNNs, which builds a bridge between tensor compression and model pruning, and achieve good results. Experiments show that QSFM can compress CNNs such as VGGNet, ResNet and MobileNet significantly with negligible accuracy drop. CNNs compressed by QSFM have faster inference speed and occupy less memory, which are more appropriate to AI on edge tasks.

    QSFM can be further promoted by using more effective functions or even some machine learning methods in the process of quantifying similarity. In addition, how to divide different kinds of similar feature maps, further refine the method of deleting feature maps and determine the distribution of feature maps suitable for CNNs are still worth exploration. We will focus on selecting better ways to measure the similarity of feature maps to improve the performance of QSFM, trying to refine the compression rate for each model, and combining various compression methods to further accelerate the AI edge device inference speed.

\bibliographystyle{IEEEtran}

\bibliography{reference}

\end{document}